\documentclass[10pt,twocolumn,letterpaper]{article}

\usepackage{cvpr}              % To produce the CAMERA-READY version

\usepackage{capt-of}
\usepackage[dvipsnames]{xcolor}
\usepackage{xspace}
\usepackage{enumitem}
\usepackage{amsmath,amssymb,amsbsy,amsfonts,dsfont,pifont,bm,bbm,mathrsfs,mathtools,nicefrac}
\usepackage{algorithm,algpseudocode,listings}
\usepackage{booktabs,multirow,adjustbox,diagbox,threeparttable,tabularray}
\usepackage{bm}

\newcommand{\xq}{\bm{X}_1}
\newcommand{\xt}{\bm{X}_t}
\newcommand{\xtm}{\bm{X}_t^m}
\newcommand{\Ftm}{\bm{F}_t^m}
\newcommand{\Ftone}{\bm{F}_t^0}

\newcommand{\Ctm}{\bm{C}_t^m}
\newcommand{\Gtm}{\bm{G}_t^m}
\newcommand{\Gitm}{\bm{G}_{i,t}^m}
\newcommand{\Rthree}{\mathbb R^3}

\newcommand{\ts}{T_s}
\newcommand{\vit}{v_{i,t}}
\newcommand{\FitM}{\bm{F}_{i,t}^M}
\newcommand{\xitM}{\bm{X}_{i,t}^M}

\newcommand{\Gm}{\mathcal{G}^{m}}
\newcommand{\Gmplusone}{\mathcal{G}^{m+1}}
\newcommand{\Eim}{\bm{E}_i^m}
\newcommand{\Ejm}{\bm{E}_j^m}

\newcommand{\smij}{s^m_{ij}}
\newcommand{\xqi}{\bm{X}_{i,1}}
\newcommand{\xqj}{\bm{X}_{j,1}}
\newcommand{\xitm}{\bm{X}_{i,t}^m}
\newcommand{\xjtm}{\bm{X}_{j,t}^m}
\newcommand{\xitmhat}{\bm{\hat{X}}_{i,t}^m}
\newcommand{\vithat}{\hat{v}_{i,t}}
\newcommand{\deltaavg}{<\hspace{-0.2em}\delta_{\text{avg}}^{x}}

\newcommand{\tocite}[1]{{\color{red} [TO CITE]}}
\newcommand{\sd}[2]{{\textcolor{cyan}{[Sida: (#1) \textcolor{gray}{#2}]}}}
\newcommand{\qw}[1]{{\color{blue} [Qianqian: #1]}}

\newcommand{\yj}[1]{{\color{cyan} [Yujun: #1]}}
\newcommand\blfootnote[1]{%
  \begingroup
  \renewcommand\thefootnote{}\footnote{#1}%
  \addtocounter{footnote}{-1}%
  \endgroup
}

\definecolor{cvprblue}{rgb}{0.21,0.49,0.74}
\usepackage[pagebackref,breaklinks,colorlinks,citecolor=cvprblue]{hyperref}
\usepackage{wrapfig}
\usepackage[capitalize]{cleveref}  % Should be loaded after 'hyperref', and works perfectly with 'subfigure'.
\crefname{section}{Sec.}{Secs.}
\Crefname{section}{Section}{Sections}
\crefname{table}{Tab.}{Tabs.}
\Crefname{table}{Table}{Tables}
\crefname{figure}{Fig.}{Figs.}
\Crefname{figure}{Figure}{Figures}
\crefname{equation}{Eq.}{Eqs.}
\Crefname{equation}{Equation}{Equations}
\def\paperID{3369}
\def\confName{CVPR}
\def\confYear{2024}

\begin{document}

% \title{Lifting High-fidelity Image Translation to Video with Implicit Deformable Representations}
% \title{Disentangled Video Representation with Hash Deformation Field}
% \title{Hash Deformation Field: An Image-friendly Video Representation}
% \title{Lifting Image Algorithms to Videos with Content Deformation Fields}
% \title{SpatialTracker: Track Any Pixels in 3D with Any Pairs Rigid Affinity}
% \title{Comet: Tracking Occluded Points in Videos via 3D Trajectory Estimation}
\title{SpatialTracker: Tracking Any 2D Pixels in 3D Space}

\author{
    Yuxi Xiao\textsuperscript{1,3*} \quad
    Qianqian Wang\textsuperscript{2*} \quad
    Shangzhan Zhang\textsuperscript{1,3} \quad
    Nan Xue\textsuperscript{3} \\[2pt]
    Sida Peng\textsuperscript{1} \quad
    Yujun Shen\textsuperscript{3} \quad
    Xiaowei Zhou\textsuperscript{1}$^{\dagger}$ \quad
    \\[5pt]
    $^1$Zhejiang University \qquad
    $^2$UC Berkeley \qquad
    $^3$Ant Group \\[8pt]
}

\twocolumn[{
\renewcommand\twocolumn[1][]{#1}
\maketitle
\begin{center}
    \vspace{-21pt}
    \includegraphics[width=1.0\linewidth]{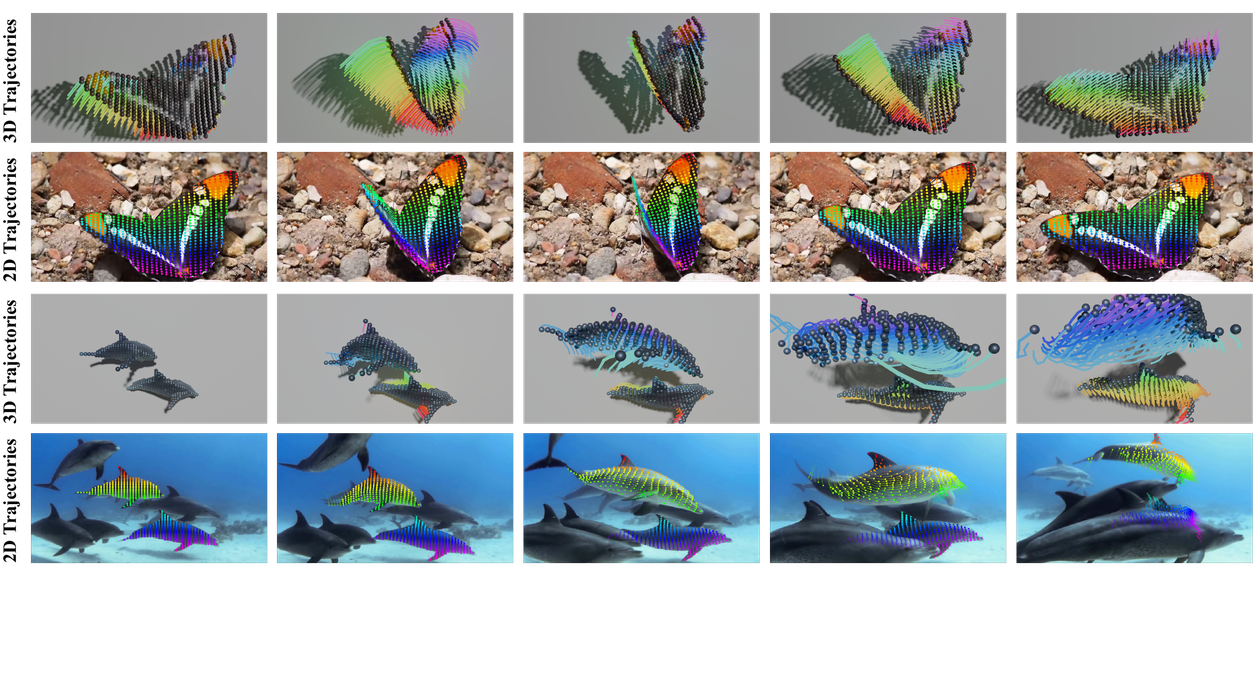}
    \vspace{-50pt}
    \captionsetup{type=figure}
    \caption{% 
        {\bf Tracking 2D pixels in 3D space.} 
        To estimate 2D motion under the occlusion and complex 3D motion, we lift 2D pixels into 3D and perform tracking in the 3D space. Two cases of the estimated 3D and 2D trajectories of a waving butterfly (top) and a group of swimming dolphins (bottom) are illustrated.
        % The first and third rows show the 3D trajectories of a waving butterfly and a group of swimming dolphins. And the second and forth rows are their 2D tracking respectively.  
        % The 3D motion of the two wings can be approximated by rotations of rigid parts and satisfy the as-rigid-as-possible constraint.
        % The 3D motion of the two wings can be approximated by two rigid groups, adhering to the as-rigid-as-possible constraint.
        % The three rows show the 3D trajectories, 2D trajectories, and rigid part labels predicted by our model, respectively.
        % The 2D motion field of a waving butterfly is complex, while its 3D motion can be approximated by rotations of rigid parts which are easier to track with the as-rigid-as-possible constraint. The three rows show the 3D trajectories, 2D trajectories, and rigid part labels predicted by our model, respectively.
    }
    \vspace{-4mm}
    \label{fig:teaser}
\end{center}
}]

\blfootnote{$^*$The first two authors contributed equally. The authors from Zhejiang University are affiliated with the State Key Lab of CAD\&CG. $^\dagger$Corresponding author: Xiaowei Zhou.}

\begin{abstract}
\vspace{-0.5em}

Recovering dense and long-range pixel motion in videos is a challenging problem. Part of the difficulty arises from the 3D-to-2D projection process, leading to occlusions and discontinuities in the 2D motion domain. While 2D motion can be intricate, we posit that the underlying 3D motion can often be simple and low-dimensional. In this work, we propose to estimate point trajectories in 3D space to mitigate the issues caused by image projection. Our method, named \emph{SpatialTracker}, lifts 2D pixels to 3D using monocular depth estimators, represents the 3D content of each frame efficiently using a triplane representation, and performs iterative updates using a transformer to estimate 3D trajectories. Tracking in 3D allows us to leverage as-rigid-as-possible (ARAP) constraints while simultaneously learning a rigidity embedding that clusters pixels into different rigid parts. Extensive evaluation shows that our approach achieves state-of-the-art tracking performance both qualitatively and quantitatively, particularly in challenging scenarios such as out-of-plane rotation. And our project page is available at \url{https://henry123-boy.github.io/SpaTracker/}.

\vspace{-4mm}
\end{abstract}
% \vspace{-10pt}

\section{Introduction}\label{sec:intro}

Motion estimation has historically been approached through two main paradigms: feature tracking~\cite{lucas1981iterative,shi1994good,tomasi1991detection,lowe2004distinctive} and optical flow~\cite{horn1981determining,teed2020raft,beauchemin1995computation}. While each type of method enables numerous applications, neither of them fully captures the motion in a video: optical flow only produces motion for adjacent frames, whereas feature tracking only tracks sparse pixels.

An ideal solution would involve the ability to estimate both \emph{dense} and \emph{long-range} pixel trajectories in a video sequence~\cite{PartVid,rubinstein2012towards}. Seminal work like Particle Video~\cite{PartVid} has bridged the gap by representing video motion using a set of semi-dense and long-range particles. More recently, several efforts~\cite{pips,tap-vid,Tapir,cotracker} have revisited this problem, formulating it as \emph{tracking any point} and addressing it through supervised learning frameworks. While trained solely on synthetic datasets~\cite{kubric,zheng2023pointodyssey}, these methods consistently demonstrate strong generalization abilities to real-world videos, pushing the boundaries of long-range pixel tracking through occlusions.

While great progress has been achieved, current solutions still struggle in challenging scenarios, particularly in cases of complex deformation accompanied by frequent self-occlusions. We argue that one potential cause for this difficulty stems from tracking only in the 2D image space, thereby disregarding the inherent 3D nature of motion. As motion takes place in 3D space, certain properties can only be adequately expressed through 3D representations. For example, rotation can be succinctly explained by three parameters in 3D, and occlusion can be simply expressed with z-buffering, but they are much more complicated to express within a 2D representation. In addition, the key component of these methods --- using \emph{2D} feature correlation to predict motion updates --- can be insufficient. Image projection
can bring spatially distant regions into proximity within the 2D space, which can cause the local 2D neighborhood for correlation to potentially contain irrelevant context~(especially near occlusion boundaries), thereby leading to difficulties in reasoning.

To tackle these challenges, we propose to leverage geometric priors from state-of-the-art monocular depth estimators~\cite{ZoeDepth} to lift 2D pixels into 3D, and perform tracking in the 3D space. This involves conducting feature correlation in 3D, which provides more meaningful 3D context for tracking especially in cases of complex motion.
Tracking in 3D also allows for enforcing 3D motion priors~\cite{raft3D,quiroga2014dense} such as ARAP constraint.
Encouraging the model to learn which points move rigidly together can help track ambiguous or occluded pixels, as their motion can then be inferred using neighboring unambiguous and visible regions within the same rigid group.

Specifically, we propose to represent the 3D scene of each frame with triplane feature maps~\cite{chan2022efficient}, which are obtained by first lifting image features to 3D featured point clouds and then splatting them onto three orthogonal planes.
The triplane representation is compact and regular, suitable for our learning framework.
Moreover, it covers the 3D space densely, enabling us to extract the feature vectors of any 3D point for tracking. We then compute 3D trajectories for query pixels through iterative updates predicted by a transformer using features from our triplane representation.
% To train our model, we utilize ground-truth 3D trajectory annotations from the Kubric dataset~\cite{kubric}. 
To regularize the estimated 3D trajectories with 3D motion prior, our model additionally predicts a rigidity embedding for each trajectory, which allows us to softly group pixels exhibiting the same rigid body motion and enforce an ARAP regularization for each rigid cluster.
We demonstrate that the rigidity embedding can be learned self-supervisedly and produce reasonable segmentation of different rigid parts.

Our model achieves state-of-the-art performance on various public tracking benchmarks including TAP-Vid~\cite{tap-vid}, BADJA~\cite{BADJA} and 
PointOdyssey~\cite{zheng2023pointodyssey}. Qualitative results on challenging Internet videos also demonstrate the superior capability of our model to handle fast complex motion and extended occlusion.

\begin{figure*}[!t]
    \centering
    \includegraphics[width=0.95\linewidth]{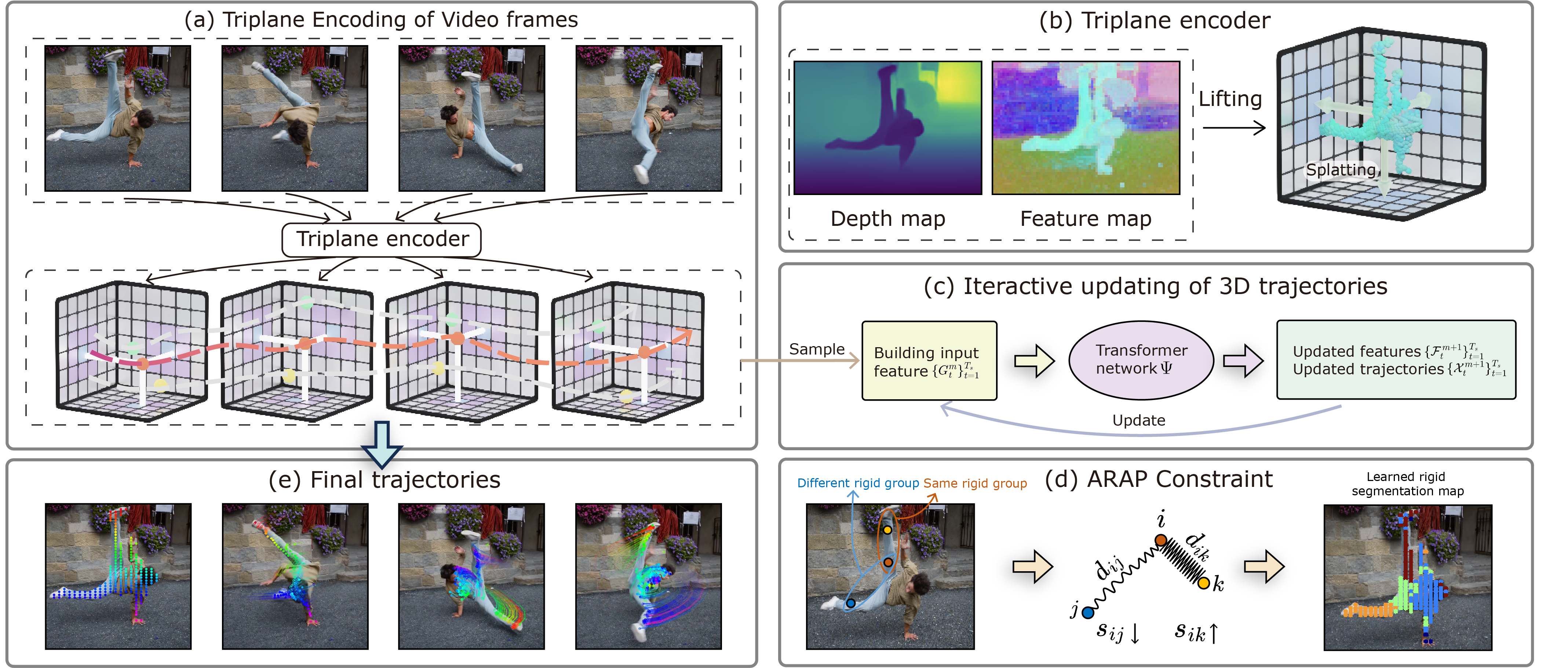}
    % \vspace{-10pt}
    \caption{\textbf{Overview of Our Pipeline.} We first encode each frame into a triplane representation (\textbf{a}) using a triplane encoder (\textbf{b}). We then initialize and iteratively update point trajectories in the 3D space using a transformer with features extracted from these triplanes as input (\textbf{c}). The 3D trajectories are trained with ground truth annotations and are regularized by an as-rigid-as-possible (ARAP) constraint with learned rigidity embedding (\textbf{d}). The ARAP constraint enforces that 3D distances between points with similar rigidity embeddings remain constant over time. Here $d_{ij}$ represents the distance between points $i$ and $j$, while $s_{ij}$ denotes the rigid similarity. Our method produces accurate long-range motion tracks even under fast movements and severe occlusion (\textbf{e}). }
    \vspace{-4mm}
    \label{fig:pipeline}
\end{figure*}

\section{Related Work}\label{sec:related}

\paragraph{Optical flow.}
Optical flow is the task of estimating dense 2D pixel-level motion between a pair of frames. While traditional methods~\cite{black1993framework,beauchemin1995computation,horn1981determining,brox2009large,weinzaepfel2013deepflow,zach2007duality} formulate it as an energy minimization problem, recent approaches~\cite{fischer2015flownet,ilg2017flownet,sun2018pwc,hui2018liteflownet,xu2017accurate} have demonstrated the ability to predict optical flow directly using deep neural networks. Notably, RAFT~\cite{teed2020raft} employs a 4D correlation volume and estimates optical flow through iteratively updates with a recurrent operator. More recently, transformer-based flow estimators~\cite{huang2022flowformer,shi2023flowformer++,zhao2022global,jaegle2021perceiver} achieved superior performance, showing the strong capacity of the transformer architecture. However, pairwise optical flow methods are not suitable for long-term tracking, as they are not designed to handle long temporal contexts~\cite{brox2009large,weinzaepfel2013deepflow}. Multi-frame optical flow methods~\cite{ren2019fusion,janai2018unsupervised,irani1999multi,volz2011modeling,kennedy2015optical} extend pairwise flow by incorporating multi-frame contexts~(typically 3-5 frames), but this remains insufficient for tracking through long occlusions in videos spanning tens or hundreds of frames.

\vspace{-0.5em}
\paragraph{Tracking any point.} Recognizing the limitations of optical flow, seminal work Particle Video~\cite{PartVid} proposes to represent video motion as a set of long-range particles that move through time, which are optimized by enforcing long-range appearance consistency and motion coherence with variational techniques. However, Particle Video only generates semi-dense tracks and cannot recover from occlusion events~\cite{rubinstein2012towards}. Recently, PIPs~\cite{pips} revisited this idea by introducing a feedforward network that takes RGB frames of a fixed temporal window~(8 frames) as input and predicts the motion for any given query point through iterative updates. However, PIPs tracks points independently, neglecting spatial context information, and will lose the target if they stay occluded beyond the temporal window.
Several recent advancements~\cite{neoral2023mft,tap-vid,Tapir,wang2023tracking,bian2023context,zheng2023pointodyssey} in point tracking have surfaced, addressing some of PIPs' limitations.
TAPIR~\cite{Tapir} relaxes the fixed-length window constraint by using a temporal depthwise convolutional network capable of accommodating variable lengths. CoTracker~\cite{cotracker} proposed to jointly track multiple points and leverage spatial correlation between them, leading to state-of-the-art performance.

Though significant progress has been made, these works all compute feature correlation in the 2D image space, losing important information about the 3D scene where the motion actually takes place. In contrast, we lift 2D points into 3D and perform tracking in the 3D space. The more meaningful 3D contexts (as opposed to 2D), along with an as-rigid-as-possible regularization, facilitate improved handling of occlusions and enhance tracking accuracy. Previous studies have also explored computing 2D motion with a touch of 3D, e.g., through depth-separated layers~\cite{kasten2021layered,ye2022deformable,sun2010layered,sevilla2016optical} or quasi-3D space~\cite{wang2023tracking}. However, distinct from their optimization-based pipelines, we perform long-range 3D tracking in a more efficient, feedforward manner. 

\vspace{-0.5em}
\paragraph{Scene flow.}
Scene flow defines a dense 3D motion field of points in a scene. Early work estimates scene flow in multi-view stereo settings~\cite{vedula1999three,pons2007multi,zhang20013d} through variational optimization~\cite{huguet2007variational}. The introduction of depth sensors enabled more effective scene flow estimation from pairs or sequences of RGB-D frames~\cite{hadfield2011kinecting,raft3D,jaimez2015primal,wang2020flownet3d++,hornacek2014sphereflow,herbst2013rgb,quiroga2014dense}.  A considerable number of recent scene flow methods rely on stereo inputs~\cite{ma2019deep,menze2015object}, but many of them are tailored specifically for self-driving scenes, lacking generalizability to diverse non-automotive contexts. Another line of research~\cite{liu2019flownet3d,wu2020pointpwc,gu2019hplflownet,niemeyer2019occupancy,lei2022cadex,bozic2021neuraldeformationgraphs} estimates 3D motion from a pair or a sequence of point clouds. 
An important prior that is often used for scene flow estimation is local rigidity~\cite{vogel20113d,vogel2013piecewise}, where pixels are grouped into rigidly moving clusters~(object or part-level), in either a soft or hard manner. For example, RAFT-3D~\cite{raft3D} learns rigid-motion embeddings to softly group pixels into rigid objects. Scene flow estimation is also often solved as a sub-task in non-rigid reconstruction pipelines~\cite{keller2013real,lin2022occlusionfusion,newcombe2011kinectfusion}. For example, DynamicFusion~\cite{dynamicfusion} takes depth maps as input and computes dense volumetric warp functions by interpolating a sparse set of transformations as bases. In contrast to prior works, we learn to predict long-range 3D trajectories through supervised learning, providing generalization capabilities for handling complex real-world motion.

% Another line of work are non-rigid reconstruction methods like DynamicFusion, which solve 3D tracking as a sub-task using a sparse set of transformations as bases and defining the dense volumetric warp function through interpolation.

% \noindent\textbf{Implicit Neural Representations.}
% %

% \sd{We need to cite convolutional occupancy network, ALTO: Alternating Latent Topologies for Implicit 3D Reconstruction}{}

% \sd{We need to cite occupancy flow, neural deformation graph}{}

% \noindent\textbf{Consistent Video Editing.}
% %

% \noindent\textbf{Video Processing via Generative Models.}
% %

\section{Method} \label{sec:method}

% Video processing
% 1. Monocular depth estimation
% 2. Lift RGB-D video to featured Pseudo 3D points via orthographic projection

% 3D tracking module
% 1. 3D Volume Encoding with Tri plane encoder. ->> 参考 convolutional occupancy network
% 2. Low Rank SE3 Motion Basis
% 3. Track update & Temporal and Spatial Transformer
% 4. Rigid part segmentation

% Training
% 1. Loss functions

% \yx{Describe our task as the 3D tracking. And introduce the formulation and modules designs here.}{}

% \noindent\textbf{Overview.}

% Overview -> Tracking any pixels in 3D space 
% 1. Our task: input (video) and output (2D tracking, 3D tracking, and rigid part segmentation)
% 2. Our modules: video processing, 3D tracking module, training

% We formulate motion estimation from a monocular video as follows:
% \sd{Too many unused symbols, which make it difficult to read. So I remove them}{}

% Given a monocular RGB video, our goal is to estimate dense and long range 3D trajectories for arbitrary pixels.
Given a monocular video as input, our method tracks any given query pixels across the entire video. Different from prior methods that establish correspondences solely in the 2D space, we lift pixels to 3D using an off-the-shelf monocular depth estimator and perform tracking in a 3D space with richer and more spatially meaningful 3D contextual information, thereby enhancing the overall tracking performance.
% \qw{todo: should emphasize more somewhere that we only have \emph{pseudo} 3D trajectories when the input is only RGB video}
% Given a monocular RGB video $V$ comprised of $\{\mathbf{I}_1, \mathbf{I}_2, ..., \mathbf{I}_n\}$, our goal is to estimate a dense and long range 3D trajectories $\mathcal{F}_{i}$ for arbitrary pixels $\mathbf{P}_{i}\in \mathbb{R}^{2}, i=1...N$.
% For this input video, we will first use a MOE (monocular estimation) model $\mathcal{M}$ to estimate the depths map $\mathcal{M}(V)=\{\mathbf{D}_1,\mathbf{D}_2,...,\mathbf{D}_n\}$.
% Our 3D tracker algorithm $\mathcal{T}$ takes the video, depth maps and selected pixel as input and output a consistent 3D trajectory for the tracked pixels $\mathcal{T}: (V, \mathcal{M}(V), \mathbf{P}_{i}) \rightarrow \mathcal{F}_{i}$.
% Here, the $\mathcal{F}_i=\{\hat{\mathbf{P}}_{i}^{t}, \hat{\mathbf{P}}_{i}^{t}\in \mathbb{R}^{3},t=t_i, ..., T, i=0,...N\}$ is a sequence of 3D locations of tracked pixels where $t_i$ is start time of $\mathbf{P}_{i}$ tracking. 
% And to avoid the ambiguity during tracking~\tocite{}, the start tracked pixel $\mathbf{P}_{i}$ should be a physical visible point on the object surface, and its corresponded 3D point $\hat{\mathbf{P}}_{i}^{t_i}$ are attained by the querying the pixel depth from $\mathcal{M}(V)$.  

\cref{fig:pipeline} presents the overview of our proposed pipeline. 
% We follow the sliding windows fashion~\cite{pips,cotracker}, which construct the sliding windows online and propagation the trajectories with the rolling structure (Sec.~\ref{sec:Iterative}).
% And during each window, 
% We first estimate the depth for each frame using an off-the-shelf monocular depth estimator, and encode the appearance and 3D information of each frame onto a triplane representation~(Sec.~\ref{sec:fte}).
We first encode the appearance and geometry information of each frame into a triplane representation~(Sec.~\ref{sec:fte}).
Then we perform iterative prediction of trajectories in the 3D space using these triplanes in a sliding window fashion~(Sec.~\ref{sec:Iterative}). We leverage the as-rigid-as-possible (ARAP) 3D motion prior during training to facilitate tracking especially in challenging scenarios of occlusion and large motion~(Sec.~\ref{sec:arap}). 
Finally, we describe our training strategy in Sec.~\ref{sec:training}.

\subsection{Triplane Encoding of Input Video Frames}
\label{sec:fte}

To perform tracking in the 3D space, we need to lift 2D pixels into 3D and construct a 3D representation that encodes the feature for each 3D location. To this end, we propose to use triplane features as the 3D scene representation for each frame detailed below.

To start with, for each frame, we obtain its monocular depth map using a pretrained monocular depth estimator, alongside multi-scale feature maps generated by a convolutional neural network (CNN). Subsequently, 2D pixels are unprojected into a set of 3D point clouds, where each 3D point is associated with a feature vector. This feature vector is a concatenation of the corresponding image feature and a positional embedding~\cite{vaswani2017attention} of its 3D location.
 
While this featured point cloud captures both geometry and appearance information, it is incomplete and only covers visible regions (2.5D). Additionally, its irregular and unordered nature poses challenges for effective learning. One simple solution involves voxelizing the point cloud into a 3D feature volume and completing it with 3D convolutions. Yet, this approach is memory and computationally intensive. To obtain 3D features densely and efficiently, we propose to use triplane feature maps, which are obtained by orthographically projecting and average splatting~\cite{softsplat} the featured point cloud onto three orthogonal 2D planes, as illustrated in Fig.~\ref{fig:pipeline}(b).
% Specifically, 
% \qw{mention how you normalized the depth maps}  
% \yx{We simply normalize the depth map with the max and min values in a sliding windows, and}{}
% \qw{this is not clear. Did you normalize it to [0,1]? Did you compute a single min and max for the whole sliding window? We haven't introduced the sliding window yet. Maybe it's better to move it to the implementation details.}
% we orthographically project the featured point cloud on each of the triplanes and apply average splatting~\cite{softsplat} to obtain the feature map on each plane. 
Finally, additional convolutional layers are applied to process and complete each feature map. This triplane feature encoding process is applied to each video frame. Since we do not assume access to camera poses, each triplane is defined within the camera coordinate frame of its respective frame.

This triplane representation is compact and enables us to efficiently represent the 3D feature for any given 3D point within the field of view. This process involves projecting the point onto three feature planes, extracting its corresponding feature vectors through bilinear interpolation, and fusing them via simple addition.

Note that while similar concepts of triplanes are explored in related fields~\cite{peng2020convolutional,wang2023alto,chan2022efficient}, our focus here is distinct. Rather than learning a triplane to represent the 3D scene from scratch, we directly leverage monocular depth priors to obtain a triplane where the primary objective is to facilitate tracking in the 3D space, introducing a novel perspective to the field of pixel tracking.

% where the $H, W$ are the height and weight of original image and $D$ is the resolution along the $z$ axis.
% And specifically, the splatting operation (as seen in Fig.~\tocite{}) just project the dense feature points into the dual planes and aggregate them simply using the average.

% And for any location $(u, v)$ in $\mathbf{F}_{\text{yz}}$ or $\mathbf{F}_{\text{xz}}$:
% \begin{equation}
%     \begin{aligned}
%         &\mathbf{F}_{\text{yz}}(u, v) = \frac{\sum_{\mathbf{P}_a\in \Omega}\mathbf{F}_{\text{xy}}(\mathbf{P}_a)}{N}, \\
%         &\Omega = \{(u_x, v_y)|\mathcal{M}(V)(u_x, v_y)=v\},
%     \end{aligned}
% \end{equation}
% here $u=0,...\frac{H}{4}, v=0,...,D$, and $\mathbf{F}_{\text{xz}}$ is calculated in the same way. 

% The splatting is followed with two small convolutional layers to enhance the features in dual planes.

% Meanwhile, due to the huge domain gap between the training dataset and real data, we observe that the neural networks have better be only used in representative learning for image features extraction.

\subsection{Iterative Trajectory Prediction}
\label{sec:Iterative}
Given a set of query pixels in the query frame, Sec.~\ref{sec:fte} allows us to obtain their 3D locations and their corresponding triplane features. We now describe the process to estimate their 3D trajectories across the entire video. 

Following CoTracker~\cite{cotracker}, we partition the video into overlapping windows of length $\ts$. In each window, we iteratively estimate 3D trajectories for query points over $M$ steps using a transformer.  The final 3D trajectories are then propagated to the next window and updated, and this process continues until the end of the video.

\vspace{-1em}
\paragraph{Iterative prediction.} We now focus on the iterative prediction of 3D trajectories within the first temporal window. Given the 3D location $\xq\in\Rthree$ of a query pixel in the first frame, our goal is to predict its 3D corresponding locations~(or in other words, its 3D trajectory) in subsequent frames $\{\xt\}_{t=2}^{\ts}$, where $t$ is the frame index.

Because we adopt an iterative updating strategy to estimate the 3D trajectories, we further denote the prediction at the $m$-th step as $\{\xtm\}_{t=2}^{\ts}$. To start with, we initialize $\{\xt^0\}_{t=2}^{\ts}$ to be all equal to $\xq$, and then we iteratively update the 3D trajectory using a transformer $\Psi$.

Specifically, for the point $\xtm$ at the $m$-th iteration, we define its input feature $\Gtm$ to the transformer as:
\begin{equation}
    \Gtm = [\gamma(\xtm), \Ftm, \Ctm, \gamma(\xtm - \xq)] \in \mathbb{R}^D,
    \label{eq:input_feature}
\end{equation}
where $\gamma$ is the positional encoding function and $\Ftm$ is the feature of point $\xtm$. At the first iteration, $\Ftone$ is extracted from the triplane of frame $t$ at $\xt^0$.
For later iterations, $\Ftm$ is a direct output of the transformer from the previous iteration. $\Ctm$ denotes correlation features, which are computed by comparing $\Ftm$ and local triplane features around $\xtm$ at frame $t$.
More details of correlation features can be found in the supplementary material.

For each update, the transformer takes as input the features for the trajectories of all query points across the entire window. We denote this set of features at the $m$-th iteration as $\Gm \in \mathbb{R}^{N\times\ts\times D}  = \{\Gitm \ | \ i=1, ..., N; t=1, ..., T_s\}$, where $i$ is the query point index and $N$ is the number of query points. $\Psi$ then takes $\Gm$ as input and predicts the new set of point positions and features:
\begin{equation}
    \mathcal{X}^{m+1}, \mathcal{F}^{m+1} = \Psi(\Gm), 
\end{equation}
where $\mathcal{X}^{m+1}$ denotes the set of updated point positions, and $\mathcal{F}^{m+1}$ denotes the set of updated point features. New $\Gmplusone$ can then be defined according to Eq.~\ref{eq:input_feature}, and the same process is repeated $M$ times to obtain the final 3D trajectories for all query points $\mathcal{X}^M=\{\bm{X}_{i,t}^M\}$.
% Based on $\mathcal{X}^{m+1}$ and $\mathcal{F}^{m+1}$, we construct the set of track features $\mathcal{G}^{m+1}$ according to Eq.~\eqref{eq:track_feature}.
% Our approach iteratively updates 3D trajectories by $M$ times.
The 2D correspondence predictions can be computed by simply projecting $\{\xitM\}$ back onto the 2D image plane.
% The specific network architecture can be found in the supplementary material.

As query pixels may not have corresponding pixels at some frames due to occlusions, we additionally predict the visibility for each point of the 3D trajectories at the final iteration $M$. Specifically, for each point $\xitM$, we employ an MLP network that takes the feature $\FitM$ as input and predicts a visibility score $\vit$.

\paragraph{Handling long videos.}
% 1. motivation：为了处理长视频，构造sliding windows
% 2. 怎么根据上一个sliding window初始化下一个sliding window
To track points across a long video, we utilize overlapping sliding windows where each pair of adjacent windows has half of their frames overlapped.
Given the results from the previous window, we initialize trajectories of the first $\frac{\ts}{2}$ frames of the current window by copying the results of the last $\frac{\ts}{2}$ frames from the previous window.
The trajectories of the last $\frac{\ts}{2}$ frames in the current window are initialized by copying the result of the frame $\frac{\ts}{2}$.

\subsection{As Rigid As Possible Constraint}
\label{sec:arap}

% Motivation: 3D有约束
% 具体做法：预测每个trajectory的embedding，然后算N个query pixels的affinity matrix
% 算Loss：distance、affinity matrix

An advantage of tracking points in 3D is that we can enforce an as-rigid-as-possible~(ARAP) constraint, which enhances spatial consistency and facilitates the prediction of motion especially during occlusions.
% 1. 给定T x N x D的point features，沿着temporal dimension做averaging，并对每个feature做normalization，得到N x D的aggregated features
% 2. 对aggregated features做内积，得到affinity matrix

Enforcing proper ARAP constraints requires identifying if two points belong to the same rigid part.
To this end, at each iteration $m$, we additionally compute a rigidity embedding $\Eim$ for each trajectory by aggregating its features $\{\Gitm\}_{t=1}^{T_s}$ across time. Then, the rigidity affinity $\smij$ between any two trajectories $i$ and $j$ can be calculated as:
\begin{equation}
    \smij = \text{sim}(\Eim, \Ejm),
\end{equation}
where $\text{sim}(\cdot, \cdot)$ is the cosine similarity measure.

By the definition of rigidity, the distances between points that are rigidly moving together should be preserved over time. Therefore, we formulate our ARAP loss as follows, encouraging the distances between pairs of points exhibiting high rigidity to remain constant over time:
\begin{equation}
\resizebox{0.95\hsize}{!}{
    $\mathcal{L}_{\text{arap}} = \sum\limits_{m=1}^M
    \sum\limits_{t=1}^{T_s}
    \sum\limits_{\Omega_{ij}} w^m
    \smij ||d(\xitm, \xjtm) - d(\xqi, \xqj)||_1$,
\label{eq:arap}

}
\end{equation}
where $\Omega_{ij}$ is the set of all pairwise indices and $d(\cdot, \cdot)$ is the Euclidean distance function, and $w^m = 0.8^{M-m}$ is the weight for the $m$-th step. This ARAP loss provides gradients for learning both the 3D trajectories and the rigidity embeddings.

Based on the affinity score between any two points, we can perform spectral clustering~\cite{von2007tutorial,scikit-learn} to obtain the segmentation of query pixels.
Experiments in Sec.~\ref{sec:exp} show that our method can generate meaningful segmentation of rigid parts.

\subsection{Training}
\label{sec:training}
% 1. Overview
% 2. 描述各个objective function
% 3. The total loss
% 4. unrolled learning

In addition to the ARAP loss, we supervise the predicted trajectories using ground truth 3D trajectories at each iteration, which is defined as:
\begin{equation}
    \mathcal{L}_{\text{traj}} = \sum_{m=1}^{M} \sum_{i=1}^{N} \sum_{t=1}^{T_s} w^m ||\xitm -\xitmhat||_1,
\end{equation}
where $\xitm$ and $\xitmhat$ are the predicted and ground-truth 3D corresponding locations, respectively, and $w^m$ is the weight for the $m$-th step, identical to that in Eq.~\ref{eq:arap}.

The predicted visibilities are supervised using:
\begin{equation}
    \mathcal{L}_{\text{vis}} = \sum_{i=1}^{N}\sum_{t=1}^{T_s} \text{CE}(\vit, \vithat),
\end{equation}
where $\vit$ and $\vithat$ denote the predicted and ground-truth visibility, respectively.
$\text{CE}$ represents the cross entropy loss.

The total loss function for training is defined as:
\begin{equation}
    \mathcal{L}_{\text{total}} = \mathcal{L}_{\text{traj}} + \alpha\mathcal{L}_{\text{vis}} + \beta\mathcal{L}_{\text{arap}},
\end{equation}
where $\alpha$ and $\beta$ are weighting coefficients.
In practice, they are set as $10$ and $0.1$, respectively.

% \qw{unclear what this sentence means}
% We follow the unrolled learning strategy proposed in CoTracker~\cite{cotracker} to supervise the semi-overlapped sliding windows.

\subsection{Implementation Details}
We train our model on the TAP-Vid-Kubric dataset~\cite{tap-vid,kubric}. Our training data contains 11,000 24-frame RGBD sequences with full-length 3D trajectory annotations. During training, we use ground truth depth maps and camera intrinsics to unproject pixels into 3D space. In cases where the depth map and intrinsics are unavailable at inference, we use ZoeDepth~\cite{ZoeDepth} to predict the metric depth map for each video frame, and simply set the focal length to be the same as the image width. To generate triplane feature maps, we discretize the depth values into $d=256$ bins.
The resolutions of the triplane feature maps are $h\times w$, $w\times d$, $h\times d$ for XY, XZ, and YZ planes, respectively, where $h, w$ are the image height and width. The number of channels of the triplane features is $128$.

% For both training and inference, the depth maps are normalized into $[0, 1]$ and split into 256 bins for triplane splatting 

\begin{table*}[t]
\vspace{-8pt}
\centering\small
\SetTblrInner{rowsep=1.0pt}      % Row space.
\SetTblrInner{colsep=6pt}      % Col space.
\begin{tblr}{
    cells={halign=c,valign=m},   % Text alignment for all cells.
    column{1}={halign=l},        % Text alignment for the first role.
    cell{1}{1}={r=2}{},          % Multi-row cells.
    cell{1}{2,5,8,11}={c=3}{},   % Multi-column cells.
    hline{1,3,9}={1-13}{},       % Horizontal lines.
    hline{2}={2-13}{},           % Horizontal lines.
    hline{1,9}={1.0pt},          % Horizontal line width.
    vline{2,5,8,11}={1-8}{},     % Vertical lines.
}
    \textbf{Methods} & 
    \textbf{Kinetics}~\cite{kinetics} & & & \textbf{DAVIS}~\cite{davis} & & & \textbf{RGB-Stacking}~\cite{rgb-stacking} & & & \textbf{Average} & & \\
    & $\text{AJ}\uparrow$ & $<\delta_{\text{avg}}\uparrow$ & $\text{OA}\uparrow$ & $\text{AJ}\uparrow$ & $<\delta_{\text{avg}}\uparrow$ & $\text{OA}\uparrow$ & $\text{AJ}\uparrow$ & $<\delta_{\text{avg}}\uparrow$ & $\text{OA}\uparrow$& $\text{AJ}\uparrow$ & $<\delta_{\text{avg}}\uparrow$ & $\text{OA}\uparrow$ \\
    TAP-Net~\cite{tap-vid}     & $38.5$ & $54.4$ & $80.6$ & $33.0$ & $48.6$ & $78.8$ & $54.6$ & $68.3$ & $87.7$ & $42.0$ & $57.1$ & $82.4$\\
    PIPs~\cite{pips}           & $31.7$ & $53.7$ & $72.9$ & $42.2$ & $64.8$ & $77.7$ & $15.7$ & $28.4$ & $77.1$ & $29.9$ & $50.0$ & $75.9$\\
    OmniMotion~\cite{wang2023tracking} &  & - & - & $46.4$ & $62.7$ & $85.3$ & $\mathbf{69.5}$ & $\mathbf{82.5}$ & $\mathbf{90.3}$ & - & - & -\\
    TAPIR~\cite{Tapir}         & $49.6$ & $64.2$ & $85.0$ & $56.2$ & $70.0$ & $86.5$ & $54.2$ & $69.8$ & $84.4$ & $53.3$ & $68.0$ & $85.3$\\
    CoTracker~\cite{cotracker} & $48.7$ & $64.3$ & $86.5$ & $60.6$ & $75.4$ & $89.3$ & $63.1$ & $77.0$ & $87.8$ & $57.4$ & $72.2$ & $87.8$ \\
    Ours & $\mathbf{50.1}$ & $\mathbf{65.9}$ & $\mathbf{86.9}$ & $\mathbf{61.1}$ & $\mathbf{76.3}$ & $\mathbf{89.5}$ & 
    ${63.5}$ & 
    ${77.6}$ & 
    ${88.2}$ & 
    $\mathbf{58.2}$ & $\mathbf{73.3}$ & $\mathbf{88.2}$ \\
\end{tblr}
\vspace{-5pt}

\iffalse
\resizebox{\linewidth}{!}{ 
\begin{tabular}{l|ccc|ccc|ccc|ccc}
\toprule
\multirow{2}{*}{Method} & \multicolumn{3}{c|}{Kinetics~\cite{kinetics}} & \multicolumn{3}{c|}{DAVIS~\cite{davis}} & \multicolumn{3}{c|}{RGB-Stacking~\cite{rgb-stacking}} & \multicolumn{3}{c}{\textbf{Average}} \\
% \cmidrule{2-13}
  & $\text{AJ}\uparrow$ & $<\delta_{\text{avg}}\uparrow$ & $\text{OA}\uparrow$ & $\text{AJ}\uparrow$ & $<\delta_{\text{avg}}\uparrow$ & $\text{OA}\uparrow$ & $\text{AJ}\uparrow$ & $<\delta_{\text{avg}}\uparrow$ & $\text{OA}\uparrow$& $\text{AJ}\uparrow$ & $<\delta_{\text{avg}}\uparrow$ & $\text{OA}\uparrow$ \\
\midrule
 TAP-Net~\cite{tap-vid}&38.5&54.4&80.6&33.0&48.6&78.8&-&-&-&-&-& -\\
 PIPs~\cite{pips} &31.7&53.7&72.9&42.2&64.8&77.7&-&-&-&-&-&-\\
 TAPIR~\cite{Tapir}&48.7&64.2&85.0&56.2&70.0&86.5&-&-&-&-&-&-\\
 Cotracker~\cite{cotracker}&48.7&64.3&86.5&58.8&74.6&88.5&63.1&77.0&87.8&56.9&71.9&87.6 \\
 Ours&$\textbf{50.1}$&$\textbf{65.9}$&$\textbf{86.9}$&$\textbf{59.6}$&$\textbf{75.7}$&$\textbf{88.9}$&$\textbf{63.5}$&$\textbf{77.6}$&$\textbf{88.2}$&$\textbf{57.7}$&$\textbf{73.1}$&$\textbf{88.0}$\\
 
\bottomrule
\end{tabular}%
}
\fi

\caption{\textbf{2D Tracking Results on the TAP-Vid Benchmark.} We report the average jaccard~(AJ), average position accuracy~($\deltaavg$), and occlusion accuracy (OA) on Kinetics~\cite{kinetics}, DAVIS~\cite{davis} and RGB-Stacking~\cite{rgb-stacking} datasets.}
\vspace{-2mm}
\label{table:Tracking2D}
\end{table*}

% And we clamp the depth map with the near and far distance of 0.1 and 65 meters.
% And we normalize the depth map for each sequence by their max-min to normalize the depth into [0,1]. And in triplane encoding, the z axis of depth are split into 256 bins for triplane splating and sampling. Besides,
We train our model with eight 80GB A100 GPUs for 200k iterations. The total training time is around 6 days. The iteration steps $M$ and sliding window length $T_s$ are set to $6$ and $8$ respectively. In each training batch, we sample $N=256$ query points. The transformer $\Psi$ consists of six blocks, each comprising both spatial and temporal attention layers. For more details, please refer to the supplementary material.

\section{Experiments}\label{sec:exp}

At inference, our method can operate in two different modalities. The first modality (and the primary focus of this paper) is long-range 2D pixel tracking. In this modality, the input is an RGB video without known depth or camera intrinsics, and we rely on ZoeDepth~\cite{ZoeDepth} to estimate the depth maps. Due to the lack of precise depth and intrinsics information, we only evaluate the 2D projection of the 3D trajectories onto the image plane, i.e., 2D pixel trajectories. When RGBD videos and camera intrinsics are available, our method can be used in the second modality to predict long-range 3D trajectories.
We evaluate our method for both 2D and 3D tracking performance in Sec~\ref{sec:2d tracking eval} and Sec~\ref{sec:3d tracking eval}, respectively, and then conduct ablation studies in Sec.~\ref{sec: ablation}.

% For 2D tracking, we evaluate on three datasets from the TAP-Vid benchmark~\cite{tap-vid} which are DAVIS~\cite{davis}, Kinetics\cite{kinetics} and RGB-Stacking~\cite{rgb-stacking}, as well as BADJA~\cite{BADJA} and PointOdyssey~\cite{zheng2023pointodyssey}. For the evaluation of 3D tracking in RGBD videos, we evaluate our method on PointOdyssey\cite{zheng2023pointodyssey}, where we take ground-truth depth and intrinsics as input to our method.

% Our model is trained on the synthetic dataset, kubric~\cite{kubric}, where we take the RGBD data as input and supervise the model with the generated 3D point trajectories. For evaluation of 2D tracking performance, we conduct experiments on the real-data benchmarks: DAVIS~\cite{tap-vid}, TAP-Vid-Kinetics\cite{tap-vid}, BADJA~\cite{BADJA}, and FastCapture\cite{FastCapture}. Note that those benchmarks only provide RGB data, and we utilize the off-the-shelf depth estimator, ZoeDepth~\cite{ZoeDepth} to acquire the depth maps. For evaluation of 3D tracking in RGBD videos, we test our method on the challenging synthetic dataset, PointOdyssey\cite{zheng2023pointodyssey}, where we take ground-truth depth as input to our model and evaluate the estimated 3D trajectories.
% \qw{suggest breaking 2d and 3d tracking into two sections. Introduce in each section the datasets, metrics, and results.
% }

\subsection{2D Tracking Evaluation}
\label{sec:2d tracking eval}

We conduct our evaluation on three long-range 2D tracking benchmarks: TAP-Vid~\cite{tap-vid}, BADJA~\cite{BADJA} and 
PointOdyssey~\cite{zheng2023pointodyssey}. Our method is compared with baseline 2D tracking methods, namely TAP-Net~\cite{tap-vid}, PIPs~\cite{pips}, OmniMotion~\cite{wang2023tracking},  TAPIR~\cite{Tapir} and CoTracker~\cite{cotracker}.
The evaluation protocols and comparison results on each of the benchmarks are represented below. 

\vspace{-1em}
% We conduct our experiments in main stream 2D tracking datasets. And we compare our method with the our baseline, Cotracker~\cite{cotracker} and other current SOTA approaches.

% \noindent{\textbf{Training Dataset.}} For fair comparison, we follow our baseline, Cotracker~\cite{cotracker}, to train our model in \textbf{TAP-Vid-Kubric}~\cite{tap-vid}. \textbf{TAP-Vid-Kubric} contains 24-frame sequences rendered from the 3D scenes with dynamic objects. And the points are randomly selected from the different frames in the video, and the selected points number is set to 256.   

\paragraph{TAP-Vid Benchmark~\cite{tap-vid}}
contains a few datasets: TAP-Vid-DAVIS~\cite{davis} (30 real videos of about 34-104 frames), TAP-Vid-Kinetics~\cite{kinetics} (1144 real videos of 250 frames) and RGB-Stacking~\cite{rgb-stacking}~(50 synthetic videos of 250 frames). Each video in the benchmark is annotated with ground truth 2D trajectories and occlusions spanning the entire video duration for well-distributed points.
% These benchmarks have the annotations of the ground truth 2d trajectories and occlusions for the well-distributed points. 
We evaluate performance using the same metrics as the TAP-Vid benchmark~\cite{tap-vid}: average position accuracy~($\deltaavg$), Average Jaccard~(AJ), and Occlusion Accuracy (OA). Please refer to the supplement for more details. We follow the ``queried first'' evaluation protocol in CoTracker~\cite{cotracker}. Specifically, we use the first frame as the query frame and predict the 2D locations of query pixels from this frame in all subsequent frames. 
% ensure the queried points are input once. And we utilize the metrics same to prior methods: Occulusion Accuracy (OA), $<\delta_{\text{avg}}^{x}$ (average accuracy rate across different thresholds \ie 1,2,4,8,16) and Average Jaccard(AJ). 
The quantitative comparisons are reported in Tab.~\ref{table:Tracking2D}, which shows our method consistently outperforms all baselines except Omnimotion across all three datasets, demonstrating the benefits of tracking in the 3D space. Omnimotion also performs tracking in 3D and obtains the best results on RGB-Stacking by optimizing all frames at once, but it requires very costly test-time optimization.
% In average, our method improve the baseline from 56.9 to 57.7 in (AJ), from 71.9 to 73.1 in $(\delta_{\text{avg}})$ and from 87.6 to 88.0 in (OA). 
% The quantitative results in show that our method is more robust in those cases thanks to tracking those points in 3D space. 
\begin{table}[t]
\centering\small
\setlength{\tabcolsep}{15pt}
\begin{tabular}{lcc}
\toprule
 \textbf{Methods} & $\textbf{segA}\downarrow$ & $\bm \delta^{3\textbf{px}}\uparrow$ \\
\midrule
TAP-Net~\cite{tap-vid}  &  $54.4$ & $6.3$  \\
PIPs~\cite{pips}  & $61.9$  & $13.5$ \\
TAPIR~\cite{Tapir} & $66.9$ & $15.2$ \\
OmniMotion~\cite{wang2023tracking} & $57.2$ & $13.2$ \\
CoTracker~\cite{cotracker}  & $63.6$ & $\mathbf{18.0}$\\
Ours & $\mathbf{69.2}$ & $17.1$ \\

\bottomrule
\end{tabular}
\vspace{-5pt}
\caption{\textbf{2D Tracking Results on the BADJA Dataset~\cite{BADJA}.} The segment-based accuracy (segA) and 3px accuracy ($\delta^{\text{3px}}$) are reported.}
% \vspace{-2mm}
\label{table:BADJA}
\end{table}

\vspace{-1em}
\begin{table}[t!]
\centering\small
\setlength{\tabcolsep}{8pt}
\begin{tabular}{lccc}
\toprule
  \textbf{Methods} & \textbf{MTE}$\downarrow$ & $\mathbf{\deltaavg}$ $\uparrow$ & \textbf{Survival}$\uparrow$ \\
\midrule
TAP-Net~\cite{tap-vid} & $37.8$ & $29.2$  & $52.8$
\\ 
PIPs~\cite{zheng2023pointodyssey} & $41.0$ & $30.4$ & $67.0$ \\
CoTracker~\cite{cotracker}  & $30.5$  & $56.2$ & $76.1$  \\ 
Ours \textit{w/} ZoeDepth~\cite{ZoeDepth} & $28.3$ & $58.4$ & $\mathbf{78.6}$ \\
Ours \textit{w/} GT depth  & $\mathbf{26.6}$ & $\mathbf{64.1}$ & $78.0$\\
\bottomrule
\end{tabular}
\vspace{-5pt}
\caption{\textbf{2D Tracking Results on the PointOdyssey Dataset~\cite{zheng2023pointodyssey}.} The Median Trajectory Error (MTE), average position accuracy ($\deltaavg$), and survival rate (Survival) are reported.}
\label{table:PointOdyssey_2d}
\end{table}

\paragraph{\textbf{BADJA}~\cite{BADJA}} is a benchmark containing seven videos of moving animals with annotated keypoints. The metrics used in this benchmark include segment-based accuracy (segA) and 3px accuracy ($\delta^{\text{3px}}$). The predicted keypoint positions are deemed accurate when its distance from the ground truth keypoint is less than $0.2\sqrt{A}$, where $A$ is the summation of the area of the segmentation mask. $\delta^{\text{3px}}$ depicts the percentage of the correct keypoints whose distances from their ground truth are within three pixels. As shown in Tab.~\ref{table:BADJA}, our method demonstrates competitive performance in terms of $\delta^{\text{3px}}$ and surpasses all baseline methods by a large margin in segment-based accuracy.

% \qw{explain why $\delta^{\text{3px}}$ is worse?}

\begin{figure*}
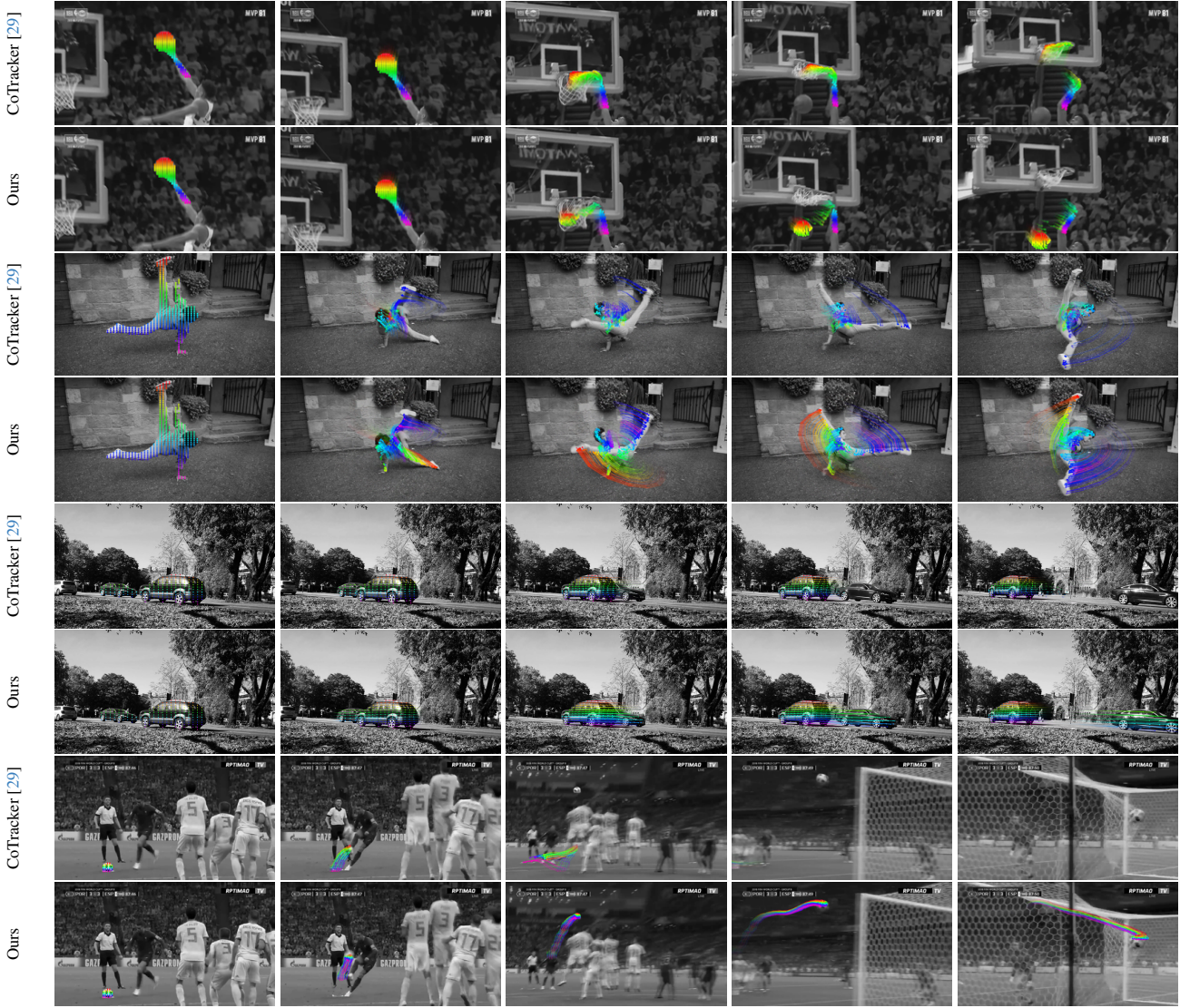

    \def\width{0.192\linewidth}
    \def\insertgraph#1#2#3{\includegraphics[width=\width]{figures/results/#1_#2-crop/frame_#3.png}}
    \def\insertgraphN#1#2#3{\includegraphics[width=\width]{figures/results/#1_#2-crop/#3.png}}
    \centering 

    \begin{minipage}[c]{0.02\linewidth}\rotatebox{90}{\scriptsize CoTracker~\cite{cotracker}}
    \end{minipage}
    \hfill
    \begin{minipage}[c]{0.96\linewidth}
    \insertgraphN{basketball}{cotracker}{00001}
    \insertgraphN{basketball}{cotracker}{00015}
    \insertgraphN{basketball}{cotracker}{00040}
    \insertgraphN{basketball}{cotracker}{00045}
    \insertgraphN{basketball}{cotracker}{00051}
    \end{minipage}

    \begin{minipage}[c]{0.02\linewidth}\rotatebox{90}{\scriptsize Ours \\ }\end{minipage}
    \hfill
    \begin{minipage}[c]{0.96\linewidth}
    \insertgraphN{basketball}{ours}{00001}
    \insertgraphN{basketball}{ours}{00015}
    \insertgraphN{basketball}{ours}{00040}
    \insertgraphN{basketball}{ours}{00045}
    \insertgraphN{basketball}{ours}{00051}
    \end{minipage}

    \begin{minipage}[c]{0.02\linewidth}\rotatebox{90}{\scriptsize CoTracker~\cite{cotracker}}\end{minipage}
    \hfill
    \begin{minipage}[c]{0.96\linewidth}
    \insertgraphN{break_dance1}{cotracker}{00001}
    \insertgraphN{break_dance1}{cotracker}{00060}
    \insertgraphN{break_dance1}{cotracker}{00068}
    \insertgraphN{break_dance1}{cotracker}{00072}
    \insertgraphN{break_dance1}{cotracker}{00077}
    \end{minipage}

    \begin{minipage}[c]{0.02\linewidth}\rotatebox{90}{\scriptsize Ours}\end{minipage}
    \hfill
    \begin{minipage}[c]{0.96\linewidth}
    \insertgraphN{break_dance1}{ours}{00001}
    \insertgraphN{break_dance1}{ours}{00060}
    \insertgraphN{break_dance1}{ours}{00068}
    \insertgraphN{break_dance1}{ours}{00072}
    \insertgraphN{break_dance1}{ours}{00077}
    \end{minipage}

    % \insertgraph{butterfly}{ours}{0001}
    % \insertgraph{butterfly}{ours}{0033}
    % \insertgraph{butterfly}{ours}{0058}
    % \insertgraph{butterfly}{ours}{0070}
    % \insertgraph{butterfly}{ours}{0091}

    % \insertgraph{butterfly}{cotracker}{0001}
    % \insertgraph{butterfly}{cotracker}{0033}
    % \insertgraph{butterfly}{cotracker}{0058}
    % \insertgraph{butterfly}{cotracker}{0070}
    % \insertgraph{butterfly}{cotracker}{0091}
    \begin{minipage}[c]{0.02\linewidth}\rotatebox{90}{\scriptsize CoTracker~\cite{cotracker}}\end{minipage}
    \hfill
    \begin{minipage}[c]{0.96\linewidth}
    \insertgraphN{crossroads}{cotracker}{00005}
    \insertgraphN{crossroads}{cotracker}{00012}
    \insertgraphN{crossroads}{cotracker}{00020}
    \insertgraphN{crossroads}{cotracker}{00024}
    \insertgraphN{crossroads}{cotracker}{00032}
    \end{minipage}

    \begin{minipage}[c]{0.02\linewidth}\rotatebox{90}{\scriptsize Ours}\end{minipage}
    \hfill
    \begin{minipage}[c]{0.96\linewidth}
    \insertgraphN{crossroads}{ours}{00005}
    \insertgraphN{crossroads}{ours}{00012}
    \insertgraphN{crossroads}{ours}{00020}
    \insertgraphN{crossroads}{ours}{00024}
    \insertgraphN{crossroads}{ours}{00032}
    \end{minipage}

    \begin{minipage}[c]{0.02\linewidth}\rotatebox{90}{\scriptsize CoTracker~\cite{cotracker}}\end{minipage}
    \hfill
    \begin{minipage}[c]{0.96\linewidth}
    \insertgraphN{cr7}{cotracker}{00001}
    \insertgraphN{cr7}{cotracker}{00030}
    \insertgraphN{cr7}{cotracker}{00050}
    \insertgraphN{cr7}{cotracker}{00090}
    \insertgraphN{cr7}{cotracker}{00111}
    \end{minipage}

    \begin{minipage}[c]{0.02\linewidth}\rotatebox{90}{\scriptsize Ours}\end{minipage}
    \hfill
    \begin{minipage}[c]{0.96\linewidth}
    \insertgraphN{cr7}{ours}{00001}
    \insertgraphN{cr7}{ours}{00030}
    \insertgraphN{cr7}{ours}{00050}
    \insertgraphN{cr7}{ours}{00090}
    \insertgraphN{cr7}{ours}{00111}
    \end{minipage}

    % \begin{minipage}[c]{0.02\linewidth}\rotatebox{90}{\scriptsize CoTracker~\cite{cotracker}}\end{minipage}
    % \begin{minipage}[c]{0.97\linewidth}
    % \insertgraph{car_occluded}{cotracker}{0005}
    % \insertgraph{car_occluded}{cotracker}{0020}
    % \insertgraph{car_occluded}{cotracker}{0025}
    % \insertgraph{car_occluded}{cotracker}{0027}
    % \insertgraph{car_occluded}{cotracker}{0030}
    % \end{minipage}

    % \begin{minipage}[c]{0.02\linewidth}\rotatebox{90}{\scriptsize Ours}\end{minipage}
    % \begin{minipage}[c]{0.97\linewidth}
    % \insertgraph{car_occluded}{ours}{0005}
    % \insertgraph{car_occluded}{ours}{0020}
    % \insertgraph{car_occluded}{ours}{0025}
    % \insertgraph{car_occluded}{ours}{0027}
    % \insertgraph{car_occluded}{ours}{0030}
    % \end{minipage}

    \caption{\textbf{Qualitative Comparison.} For each sequence we show tracking results of CoTracker~\cite{cotracker} and our method SpatialTracker.} 
    \vspace{-2mm}
\label{fig:exp}
\end{figure*}

% \vspace{-1em}

\paragraph{\textbf{PointOdyssey}~\cite{zheng2023pointodyssey}} is a large-scale synthetic dataset featuring diverse animated characters ranging from humans to animals, placed within diverse 3D environments.
% And the diversity comes from randomizing character appearance, motions, materials and environment lighting. 
We evaluate our method on PointOdyssey's test set which contains 12 videos with complex motion, each spanning approximately 2000 frames.
% Therefore, this dataset have the extremely long ground truth trajectories for some complex motions. For the evaluation here, 
We adopt the evaluation metrics proposed in PointOdyssey~\cite{zheng2023pointodyssey} which are designed for evaluating very long trajectories. These metrics include the Median Trajectory Error (MTE), $\deltaavg$ (consistent with TAP-Vid), and the survival rate. The survival rate is defined as the average number of frames until tracking failure over the video length. Tracking failure is identified when the L2 error exceeds 50 pixels at a resolution of $256\times256$. In Tab.~\ref{table:PointOdyssey_2d}, we report results for baseline methods as well as our method using depths from  ZoeDepth~\cite{ZoeDepth}~(default) and GT depth annotations.
Our method consistently outperforms the baselines across all metrics by a noticeable margin. In particular, we demonstrate that with access to more accurate ground truth depth, our performance can be further enhanced. This suggests the potential of our method to continue improving alongside advancements in monocular depth estimation.

\vspace{-1em}
\paragraph{\textbf{Qualitative Results.}}
% Besides quantitative comparisons on public benchmarks, 
We present qualitative comparisons with CoTracker~\cite{cotracker} on challenging videos from DAVIS~\cite{davis} and Internet footage in Fig.~\ref{fig:exp}.
Our method outperforms CoTracker in handling complex human motion with self-occlusions,  achieves a better understanding of rigid groups, and can effectively track small, rapidly moving objects even in the presence of occlusions. Please refer to the supplementary video for more results and better visualizations.
% We show more results of dense tracking on the surface of various objects, 
% \ie human body, soccer ball etc. 
% \qw{describe the qualitative results}
% The results demonstrate how the arap constraint contribute to more accurate tracking for those complex motions in the real world. 

% \qw{metric on pointodyssey is not defined here but later in the results section; should reorganize}

\subsection{3D Tracking Evaluation}
\label{sec:3d tracking eval}

Given an RGBD video (with known depth and intrinsics) as input, our method can estimate true 3D trajectories. Since no baseline method can directly be used for long-range 3D tracking, we construct our baselines by composing existing methods. Our first baseline is chained RAFT-3D~\cite{raft3D}. RAFT-3D is designed for pairwise scene flow estimation, so to obtain long-range scene flow, we chain its scene flow prediction of consecutive frames. Our second baseline is to directly lift the predicted 2D trajectories from CoTracker~\cite{cotracker} using the input depth maps. 

We evaluate our method and baselines on the PointOdyssey~\cite{zheng2023pointodyssey} dataset. We create $231$ testing sequences from the test set, each consisting of 24 frames and with a reduced frame rate set at one-fifth of the original. We use three evaluation metrics, namely $\rm{ATE}_{3D}$, $\delta_{0.1}$, and $\delta_{0.2}$. $\rm{ATE}_{3D}$ is the average trajectory error in 3D space.
$\delta_{0.1}$ and $\delta_{0.2}$ measure the percentage of points whose distances are within $0.1$m and $0.2$m from the ground truth, respectively.
\begin{table}[t!]
\centering\small
\setlength{\tabcolsep}{9pt}
\begin{tabular}{lccc}
\toprule
 \textbf{Methods} & $\mathbf{\rm ATE}_{3D}\downarrow$ & $\mathbf{\delta_{0.1}}\uparrow$ & $\mathbf{\delta_{0.2}}\uparrow$ \\
\midrule
Chained RAFT3D~\cite{raft3D}  &  0.70 & 0.12  & 0.25 \\
Lifted CoTracker~\cite{cotracker}  & 0.77  & 0.51 & 0.64  \\
Ours  & \textbf{0.22} & \textbf{0.59}  & \textbf{0.76 } \\

\bottomrule
\end{tabular}
\vspace{-5pt}
\caption{\textbf{3D Tracking Results on the PointOdyssey Dataset.} ${\rm ATE}_{3D}$ is the average trajectory error in 3D space and
$\delta_{t}$ measures the percentage of points whose distances are within $t$ (in meter) from the ground truth.}
\label{table:PointOdyssey}
\end{table}

\begin{table}[t]
\centering\small
\setlength{\tabcolsep}{8.5pt}
\begin{tabular}{lccc}
\toprule
  Methods & AJ$\uparrow$ & $\deltaavg$$\uparrow$ & OA$\uparrow$ \\
\midrule
Ours \textit{w/o} ARAP  & 55.1 & 71.6 & 87.4 \\
\midrule

Ours \textit{w/} DPT~\cite{ranftl2021vision} & 51.4 & 70.7 & 83.3 \\
Ours \textit{w/} MiDaS~\cite{birkl2023midas}  & 56.3  & 73.9 & 86.6  \\ 
Ours \textit{w/} ZoeDepth~\cite{ZoeDepth} (default) & \textbf{61.1} & \textbf{76.3} & \textbf{89.5}\\
\bottomrule
\end{tabular}
\vspace{-5pt}
\caption{\textbf{Ablation Study on the DAVIS
Dataset.} The Average Jaccard~(AJ), average position accuracy~($\deltaavg$), and Occlusion Accuracy (OA) are reported. We evaluate the effectiveness of the ARAP constraint and the influence of different monocular depth estimators (ZoeDepth~\cite{ZoeDepth}, MiDaS~\cite{birkl2023midas} and DPT~\cite{ranftl2021vision}). ``Ours \textit{w/} ZoeDepth" is the default model we use in our experiments.
}
\vspace{-4mm}
\label{table:ablatio_depth}
\end{table}

The results are shown in Tab.~\ref{table:PointOdyssey}.
Our method outperforms both ``Chained 
RAFT-3D'' and ``Lifted CoTracker'' consistently on all three metrics by a large margin. We found that RAFT-3D, trained on FlyingThings~\cite{MIFDB16},  generalizes poorly on PointOdyssey, possibly due to the fact that its dense-SE3 module is sensitive to domain gaps. In contrast, also trained on a different dataset (Kubric), our method exhibits strong generalization to PointOdyssey, affirming the efficacy of our design for 3D trajectory prediction. In addition, both baselines cannot handle occlusion and will lose track of points once they become occluded, hurting the performance significantly.

\subsection{Ablation and Analysis}
\label{sec: ablation}

\begin{figure}[t]
    \centering
 
    \includegraphics[width=0.3\linewidth]{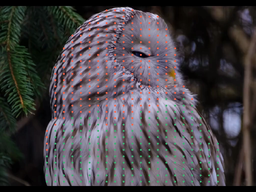}
    \includegraphics[width=0.3\linewidth]{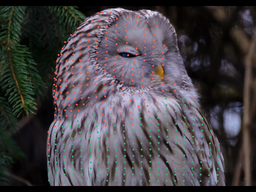}
    \includegraphics[width=0.3\linewidth]{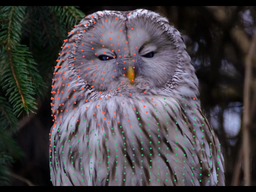}

    \includegraphics[width=0.3\linewidth]{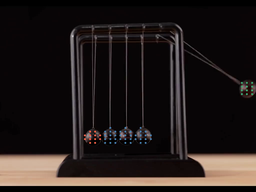}
    \includegraphics[width=0.3\linewidth]{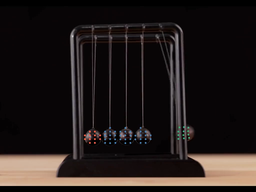}
    \includegraphics[width=0.3\linewidth]{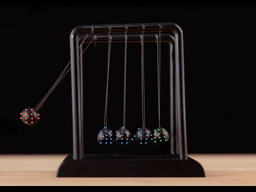}

    \vspace{-5pt}    
    \caption{\textbf{Rigid Part Segmentation.} We utilize spectral clustering on the rigidity embedding to determine rigid groups. Each color represents a distinct rigid group.}
    \vspace{-4mm}
    \label{fig:asap-rigid-seg}
\end{figure}

\paragraph{Effectiveness of ARAP loss and rigidity embedding.} We ablate the ARAP loss and report the result ``Ours \emph{w/o} ARAP'' on the TAP-Vid-DAVIS~\cite{pont20172017,tap-vid} dataset in Tab.~\ref{table:ablatio_depth}. Without the ARAP loss, the performance drops substantially, verifying the effectiveness of the ARAP constraint. We additionally showcase qualitative results of the rigid part segmentation, utilizing our learned rigidity embeddings in Fig.~\ref{fig:asap-rigid-seg}, demonstrating their effectiveness.

\vspace{-0.5em}
\paragraph{Analysis on monocular depth estimators.}
To study the influence of different monocular depth estimation methods on our model, we evaluate our method with three monocular depth models: ZoeDepth~\cite{ZoeDepth}~(default), MiDaS~\cite{birkl2023midas}, and DPT~\cite{ranftl2021vision}.
We report the results on the TAP-Vid-DAVIS~\cite{pont20172017,tap-vid} dataset in Tab.~\ref{table:ablatio_depth}.
``Ours \textit{w/} ZoeDepth'' achieves the best results, probably due to the fact that ZoeDepth~\cite{ZoeDepth} is a metric depth estimator and exhibits less temporal inconsistency across frames compared to relative depth estimators MiDaS~\cite{birkl2023midas} and DPT~\cite{ranftl2021vision}.
Furthermore, it is noteworthy that the efficacy of our model has a positive correlation with the advancements in the underlying monocular depth models. 
Please refer to the supplementary material for additional analysis and ablations.
% The results are shown in Table~\ref{table:ablatio_depth}.

% \paragraph{Effectiveness of different components.}
% To ascertain the significance of the different components of our model, we  conduct comprehensive ablation studies on the DAVIS dataset. 
% First, we design a baseline model that removes the ARAP term. This baseline is abbreviated as ``Ours w/o ARAP''.
% Second, to demonstrate the effectiveness of tracking in 3D space, we propose a baseline model that replaces the input of CoTracker with RGBD data, where the monocular depth is consistent with our model. This baseline is abbreviated as ``CoTracker w/ RGBD''.

\section{Conclusion and Discussion}\label{sec:conclusion}

In this work, we show that a properly designed 3D representation is crucial for solving the long-standing challenge of dense and long-range motion estimation in videos. Motion naturally occurs in 3D and tracking motion in 3D allows us to better leverage its regularity in 3D, e.g., the ARAP constraint. We proposed a novel framework that estimates 3D trajectories using triplane representations with a learnable ARAP constraint that identifies the rigid groups in the scene and enforces rigidity within each group. Experiments demonstrated the superior performance of our method compared to existing baselines and its applicability to challenging real-world scenarios. 

Our current model relies on off-the-shelf monocular depth estimators whose accuracy may affect the final tracking performance as shown in Tab. \ref{table:ablatio_depth}. However, we anticipate that advancements in monocular reconstruction will enhance the performance of motion estimation. We expect a closer interplay between these two problems, benefiting each other in the near future.

% \vspace{-0.5em}

\vspace{6pt}\noindent{\large\bf Acknowledgement}
\vspace{6pt} \\
\noindent This work was partially supported by National Key Research and Development Program of China (No. 2020AAA0108900), Ant Group, and Information Technology Center and State Key Lab of CAD\&CG, Zhejiang University.

{\small
\bibliographystyle{ieee_fullname}
\bibliography{main}
}

\end{document}